\documentclass[twocolumn,twoside]{IEEEtran}
\usepackage{epsfig}
\usepackage{algorithmic,algorithm}
\usepackage[utf8]{inputenc}
\usepackage{cite}
\usepackage{multirow}
\usepackage{caption}
\usepackage{amssymb}
\usepackage{bm}
\usepackage{psfrag}
\usepackage{cite}
\usepackage{graphics}
\usepackage{fancyhdr}
\usepackage{eucal}
\usepackage[english]{babel}
\usepackage{ifpdf}
\usepackage{setspace}
\usepackage{float}
\usepackage[T1]{fontenc}
\usepackage{subfigure}
\usepackage{dsfont}
\usepackage{graphicx, framed}
\usepackage{textcomp}
\usepackage{tikz}
\usepackage{tikz-inet}
\usetikzlibrary{shapes,snakes,arrows,automata}

\newif\ifnotes\notestrue

\newcommand{\ben}{\begin{enumerate}}
\newcommand{\een}{\end{enumerate}}

\newcommand{\bc}{\begin{center}}
\newcommand{\ec}{\end{center}}

\newcommand{\bit}{\begin{itemize}}
\newcommand{\eit}{\end{itemize}}

\newcommand{\ds}{\displaystyle}
\newcommand{\beq}{\begin{equation}}
\newcommand{\eeq}{\end{equation}}

\newcommand{\wre}{\mathbf{w}^{\rm{r}}}
\newcommand{\wi}{\mathbf{w}^{\rm{in}}}
\newcommand{\wo}{\mathbf{w}^{\rm{out}}}

\newcommand{\vx}{\mathbf{x}}

\newcommand{\x}{\mathbf{x}}
\newcommand{\y}{\mathbf{y}}
\newcommand{\Nx}{N_{\rm{x}}}
\newcommand{\Ny}{N_{\rm{y}}}
\newcommand{\Ns}{N_{\rm{s}}}

\newcommand{\State}{\mathbf{s}}

\newcommand{\R}{\mathds{R}}

\begin{document}
\pagestyle{empty}

\title{An Empirical Study of the L2-Boost technique with Echo State Networks}

\author{{\bf Sebasti\'an Basterrech}\\[1em]
\textit{IT4Innovations}\\
National Supercomputing Center\\
VSB-Technical University of Ostrava\\
Ostrava, Czech Republic\\
\textit{Sebastian.Basterrech.Tiscordio@vsb.cz} \\[1em]
}

\maketitle

\begin{abstract} 
%
%
%
%
%

A particular case of Recurrent Neural Network (RNN) was introduced at the beginning of the 2000s under the name of Echo State Networks (ESNs).
The ESN model overcomes the limitations during the training of the RNNs while introducing no significant disadvantages.
Although the model presents some well-identified drawbacks when the parameters are not well initialized.
The performance of an ESN is highly dependent on its internal parameters and pattern of connectivity of the hidden-hidden weights 
Often, the tuning of the network parameters can be hard and can impact in the accuracy of the models.

In this work, we investigate the performance of a specific boosting technique (called $L_2$-Boost) with ESNs as single predictors.   
The $L_2$-Boost technique has been shown to be an effective tool to combine ``weak'' predictors in regression problems.
In this study, we use an ensemble of random initialized ESNs (without control their parameters) as ``weak'' predictors of the boosting procedure.
We evaluate our approach on five well-know time-series benchmark problems.
Additionally, we compare this technique with a \textit{baseline} approach that consists of averaging the prediction of an ensemble of ESNs.
\end{abstract}


\renewcommand{\L}{\mathcal{L}}
\newcommand{\Lk}{\mathcal{L}^{(k)}}
\renewcommand{\L}{\mathcal{L}}
\newcommand{\Lm}{\mathcal{L}^{(m)}}

\section{Introduction}
\label{Introduction}

%
%
%

%
\textit{Boosting} is a general procedure for improving the accuracy of an ensemble of methods. 
It has been successful used in supervised learning problems since its apparition in the 1990s~\cite{Shapire90,Freund95,Shapire96}.
%
%
Several variations of the original Boosting idea have been introduced over the years~\cite{Friedman00, Buhlmann03}, one of the most popular is called \textit{AdaBoost}~\cite{Shapire96}. 
At the beginning, Boosting was used in problems where the output features were label or discrete responses (classification problems).  
An analogy between AdaBoost and \textit{additive models} was studied in~\cite{Friedman00}.
This connection was essential for the extension of Boosting for solving problems where the output features are continuous variables (regression problems).
B\"uhlmann \textit{et al.} developed a variation of the Boosting technique called \textit{$L_2$-Boost} that is constructed from an additive model and the functional gradient descent method~\cite{Buhlmann03}.
%
%

Since the early 2000s, a computational paradigm called \textit{Reservoir Computing (RC)} has gained prominence in the Neural Computation community.
In a RC model there are two well-separated concepts: a dynamical system and a memory-less function.
%
The purpose of the dynamical system is to encode the spatio-temporal information of the input patterns into a spatial representation.
At each time this dynamical system is characterized by its state that is called~\textit{reservoir} in the RC literature.
This non-linear transformation is most often realized by a \textit{Recurrent Neural Network (RNN)} with a large pool of interconnected neurons. 
A distinctive principle of a RC model is that the parameters of the dynamical system (the RNN weights) do not participate in the training process. 
That is, once the reservoir parameters are initialized, they remain fixed during the training process. 
%
%
Another part of the model is a memory-less supervised learning tool called \textit{readout} structure.
This part is designed to be robust and fast in the learning process.

The RC models has been applied in the neuroscience area for processing cognitive information in the neural system.~\cite{Maass10}.
Furthermore, they have proven to be extremely effective tools for time-series problems in the area of Machine Learning.
For instance, as far as we know one of the most popular RC models the \textit{Echo State Networks (ESN)}~\cite{Jaeger01}, has the best known learning performance on the Mackey-Glass time-series prediction problem~\cite{Schmidhuber07,Jaeger09}.
In this article, we will concentrate in the ESN model for solving time-series problems.
The reservoir in the ESN model is composed by a RNN with sigmoid neurons, and the readout part of the model is a linear regression.
The weight connection between neurons in the reservoir are collected in a matrix that we will call \textit{reservoir matrix}.
The main global parameters of the ESN model are: the input scaling factor, the spectral radius of the reservoir matrix and the pattern of connectivity among the reservoir units.
The setting of these parameters often requires the human expertise and several empirical trials~\cite{MantasPracticalGuide12}.
As a consequence, the setting procedure can be expensive in computational time.
For instance, the time complexity of an algorithm that computes the spectral radius of a $N\times N$ matrix is equal to $O(N^4)$~\cite{Ferreira2013}. 
%
%
%

The goal of this article is to investigate the performance of an automatic procedure to combine single \textit{weak} ESNs and the $L_2$-Boost technique.
We develop an automatic technique based on $L_2$-Boost, which combines the prediction of several random initialized ESNs in order to produce a highly accurate tool.
We use the term \textit{weak ESN} for an ESN without checking and computing the spectral radius. 
%
%
We use the terminology \textit{weak} due to the fact that this particular ESN is not optimal.
%
%
The main advantages of the procedure presented in this article are: 
\begin{itemize}
\item Descend the computational effort. In order to gain in the computational effort, the approach consists of combining weak ESNs. 
The procedure avoids to tune the reservoir parameters, which can often have a high computational cost. 
It uses only a uniform random initialization of the weights.
%
%
Note that, there are not a control of the spectral radius, then some single weak ESNs can have unstable dynamics.
%
\item The procedure is automatic. The procedure does not require external human expertise for setting the model parameters, and for evaluating the model performance. 
\item The technique has a new parameter used for \textit{overfitting control}. This parameter which we will  that comes from the $L_2$-Boost technique.
\end{itemize}

We present empirical results of the procedure introduced in this paper on a wide range of benchmark problems.
We compare these performances with the accuracy obtained by a single ESN.
Furthermore, we realize a comparison with the accuracy of a \textit{baseline} approach that computes the average among single ESN models.
Each single ESN is independently initialized and adjusted during the learning process.
There are empirical evidence in the Machine Learning literature that show that this baseline approach sometimes performs better than other ensemble methods~\cite{Opitz99}.

This work is a revised and expanded version of the article~\cite{BasterrechISDA13}.

The structure of this article is organized as follows. 
In Section~\ref{Background}, we start with a specification of supervised learning problems with temporal data.
Next, we present an overview about the family of additive models.
Subsection~\ref{L2BoostSection} introduces the $L_2$-Boost technique.
In Section~\ref{Modeling}, we present the Reservoir Computing paradigm.
Particularly, we focus on the Echo State Network model in~\ref{ESN}.
In Subsection~\ref{Methodology} is presented a formalization of the procedure introduced in this article. 
Section~\ref{Results} describes the empirical results.
This section starts with a description of the benchmark problems. 
Next, we present the reached results. 
Finally, last section provides conclusions and future work.

\newcommand{\Rx}{\R^{N_{\rm{X}}}}
\newcommand{\Ry}{\R^{N_{\rm{Y}}}}
\newcommand{\F}{\hat{F}}
\newcommand{\Fm}{\F^{(m)}}
\newcommand{\xn}{\x^{(n)}}
\newcommand{\yn}{\y^{(n)}}

\section{Background}
\label{Background}
In this Section, we start specifying the context where the ESN model and the $L_2$-Boost technique are applied.
Next, we present the additive models and we introduce a description of the $L_2$-Boost technique. 
%
%

\subsection{Problem Specification}
\label{ProblemSpecification}
We begin specifying a supervised learning problem.
Given a data set  $\L=\{(\x^{(t)},\y^{(t)}): t=1,\ldots,T\}$ where the points $\x$ and  $\y$ are either a class or a numerical response.
We denote by $\Nx$ the dimension of the input vector $\x$, and $\Ny$ the dimension of the output vector $\y$.
We suppose that the mapping between the input $\x$ and the output $\y$ is given by certain unknown function $F(\cdot)$.
The goal consists in learning a parametric function $\hat{F}(\x^{(t)},\L)$ such that certain error distance between $\F(\x^{(t)},\L)$ and $\y^{(t)}$ is minimized for all $t$.
The problem is called \textit{regression problem} when the learning set has output numerical variables.
Otherwise, it is named \textit{classification problem}.
In the case of regression problems, it is recommended to use the a quadratic distance~\cite{HastieTibshirami}. 
Even though we can also use a quadratic distance in classification problems, it is recommendable to use the Kullback-Leibler distance in this domain~\cite{HastieTibshirami}. 

An ESN model is mainly used for solving supervised learning tasks, wherein the data set presents temporal dependencies.
Although, it can be also used for non-temporal supervised learning problems~\cite{Jaeger09}.
In this article we will concentrate only in temporal learning tasks with real output variables ($\y^{(t)}\in\Ry$, for all $t$).
In this work, we perform the models using a standard discrete time.
We want to forecast some aspect of the output feature $\y$ at time $t+k$, using some aspect of the information available at current time $t$, that is given the collection $((\x^{(t)},\y^{(t)}),(\x^{(t-1)},\y^{(t-1)}),(\x^{(t-2)},\y^{(t-2)}),\ldots)$ we would like to predict the value $\y^{(t+k)}$ $(k>0)$~\cite{Bengio03}.
In this case, the goal consists in estimating a mapping $\F(\cdot)$ for predicting $\y^{(t+k)}$ for some $k>0$, such that some distance between $\F(\cdot)$ and $\y$ is minimized. 
%
%

\subsection{Additive Models}
\label{AdditiveModels}
In~\cite{Friedman00} was analyzed the Boosting model under the form of an \textit{Additive model}.
Given a set of functions \mbox{$f^{(m)}:\Rx\rightarrow \Ry$,} $m=1\ldots, M$ characterized by a set of parameters $\theta$ and expansion coefficients $\beta$,
$$
f^{(m)}(\x)=\beta^{(m)}h(\x,\theta^{(m)}),
$$
an additive model has the following form
\begin{equation}
\label{additiveModel}
F(\x)=\ds{\sum_{m=1}^{M}f^{(m)}(\x)}.
\end{equation}
The functions $\{h(x;\theta)\}^M_1$ are named \textit{basis functions}.
They are not fixed \textit{a priori} and are selected depending of the cost function used and the data set.
An important parameter of the model is the number of basis functions ($M$) considered in the expression~(\ref{additiveModel}).
This parameter controls the generalization error of the model.
Since the main goal in a learning task is to find a predictor with low generalization error, the parameter $M$ has an important role in the accuracy of an additive model.

\subsection{The $L_2$-Boost Procedure}
\label{L2BoostSection}
A relationship between the gradient descent technique and stage-wise additive expansions was introduced at the beginning of the $2000$s~\cite{Friedman00Greedy}.
The introduction of the gradient descent algorithm using a boosting approach was an essential contribution in the field of ensemble learning methods~\cite{Friedman00Greedy}.
It allowed to start to use boosting in regression problems~\cite{Buhlmann03}.
A Boost method for regression problems with quadratic error distance was introduced under the name of $L_2$-Boost in~\cite{Buhlmann03}. 
We present the $L_2$-Boost technique in Algorithm~\ref{L2Boost}. 
Other boosting variants were presented for other kind of distances, some of them are described in~\cite{Friedman00Greedy, Friedman00,Buhlmann03}.

We refer by \textit{epoch} to the iteration of the training algorithm through all the patterns in the training set~\cite{Bengio2000}.
At each epoch $m+1$, the basis function $h^{(m+1)}(\cdot,\theta)$ is fitted to the current residuals: $\y^{(i)}-\F^{(m)}(\vx^{(i)})$, for all $i$. 
Unlike other boosting techniques such as Adaboost, $L_2$-Boost does not present any re-weighting.
Another difference between $L_2$-Boost and other boosting methods is that $L_2$-Boost presents a tendency to over-fit the data~\cite{Buhlmann03}. 
The model with contracting linear learners converge to the fully saturated model~\cite{Buhlmann03}.
Each boosting epoch contributes to additional overfitting, thus the selection of the weak learners and the parameter $M$ is an essential task for this device.
In practice, few boosting iterations are enough to achieve good performances avoiding the overfitting phenomena.


\begin{algorithm}[h!t]
\caption{The $L_2$-Boost algorithm.}
   \label{L2Boost}
\begin{algorithmic}
	\REQUIRE{$\L$, $M$, $h(\x,\theta)$}
	\STATE $\;$
	\STATE Fit an initial model using a least squares fit (see~\cite{numericalRecips92}): 
		$\hat{F}^{(0)}(\cdot)=h(\cdot,\theta)$;
	\FOR{$(m=1,\ldots M)$}	
		\STATE {Compute the residuals for all pattern $i$:}
				\mbox{$\mathbf{e}^{(i)}=\y^{(i)}-\Fm(\vx^{(i)})$};
		\STATE {Fit the model $\hat{f}^{(m+1)}(\cdot)$ parametrized as}
				\mbox{$\hat{f}^{(m+1)}(\x)=h(\x,\theta)$} to the current residuals $\bm{e}$ using the least squares fit;
		\STATE {Update: $\hat{F}^{(m+1)}(\cdot)=\hat{F}^{(m)}(\cdot) + \hat{f}^{(m+1)}(\cdot)$};
	\ENDFOR	
	\STATE Return the $\hat{F}^{(M)}(\cdot)$ function;
	\end{algorithmic}
\end{algorithm}

\section{Modeling Time-series with Echo State Networks}
\label{Modeling}
The \textit{Recurrent Neural Networks (RNNs)} are powerful tools for solving time-series benchmarks.
%
They are computational methods that operate in time.
%
%
Considering terminology of graphs, in a RNN at least one circuit is presented in its topology.
The circuits of the network enable to store temporal information, in order to learn and memorize the input history~\cite{Jaeger09}.
Each circuit creates an internal state which makes the recurrent network a discrete time state-space model.
%
At each time, the RNN receives an input pattern. 
Next, the network updates its hidden state via a non-linear activation function using the input pattern and the network state at the precedent time~\cite{Martens11}.
There are a general consensus in the community that considers the RNN as powerful tool for forecasting and time-series prediction.

In spite of that, in practice the model presents some drawbacks.
The most important is that is hard to train a RNN using gradient descent methods~\cite{Bengio94}.
The training methods that use the first differential information have often stability problems and high numerical complexity.
As a consequence, much longer training times are necessary to adjust the network weights.  
In~\cite{Bengio94} is analyzed the main limitations of the algorithms of the gradient descent type for training RNNs.
%
%
These drawbacks are identified under the names of \textit{vanishing} and the \textit{exploding} gradient problems~\cite{Bengio94}.
The vanishing gradient phenomena occurs when the norm of the gradient decreases arbitrarily fast to $0$. 
The exploding gradient phenomena refers to the opposite, when the gradient norm large increases during the training process~\cite{Pascanu13}.
Recently, an effective algorithm to train RNN was introduced~\cite{Martens11}, the algorithm uses the \textit{Hessian-free Optimization} for setting the network parameters.

\textit{Reservoir Computing (RC)} models appear as a good alternative for RNNs.
The two pioneering RC models are~\textit{Echo State Network (ESN)}~\cite{Jaeger01} and \textit{Liquid State Machine (LSM)}~\cite{Maass02}.
%
%
This computational paradigm covers the main limitations related to learning processes in RNNs obtaining acceptable performance in practical applications~\cite{Jaeger09}.
In a RC model there are at least two well-differentiated structures: a dynamical system called \textit{reservoir} and another one called \textit{readout}.
The readout is a supervised learning tool for training with non-temporal data. For example: feedforward neural network, linear regression, decision trees, etc.
A main characteristic of a RC model is that the weights involved in circuits are deemed fixed during the learning process. 
%
%
Thus, the matrix with the weight between reservoir units (reservoir matrix) is initialized in an arbitrary way and it remains unchanged during the learning process.
The training algorithm is restricted to update the weights in the readout structure.
Over the last years several kinds of dynamical systems have been used for generating the reservoir state, models include: Backpropagation-decorrelation Recurrent Learning~\cite{Steil04}, Leaky Integrator Echo State Networks studied~\cite{Jaeger07}, Evolino~\cite{Schmidhuber07}, Intrinsic Plasticity~\cite{Schrauwen07}, Echo State Queueing Networks~\cite{Baster12ESQN}, Reservoir Computing and Extreme Learning~\cite{Butcher2013}, and so on.

\subsection{Formalization of the Echo State Network Model}
\label{ESN}
In this work related to the $L_2$-Boost technique and the RC methods, we only study the $L_2$-Boost with the ESN model.
An ESN reservoir is a RNN from an input space $\R^{\Nx}$ into a larger space $\R^{\Ns}$ with $\Nx \ll \Ns$. 
The connection between input and hidden neurons are collected in a~$\Ns\times\Nx$ weight matrix~$\wi$. 
The connections among the hidden neurons are represented by a~$\Ns\times\Ns$ weight matrix~$\wre$. 
A~$\Ny\times\Ns$ weight matrix~$\wo$ represents the readout weights.
At any time $t$, the information from the input pattern and the past is represented in a state vector  
\begin{equation}
\label{reservoirState}
\State(t)=\tanh (\wi \x^{(t)}+ \wre \State^{(t-1)}).
\end{equation}
%
At any time $t$, the output prediction $\y^{(t)}\in\R^{\Ny}$ is generated 
using the input pattern and the reservoir state information. Most often is computed using a linear regression:
\begin{equation}
\label{eqReadouts}
\y^{(t)}=\wo[\x^{(t)}|\State^{(t)}],
\end{equation}
where $\cdot|\cdot$ is the vertical concatenation of the vectors. 
For the sake of the notation simplicity, we omit the bias term, a constant term is included in all the regressions.
%
%

In~\cite{Jaeger01} was analyzed the stability of the reservoir dynamics in the ESN model. 
Under certain algebraic conditions the reservoir state only depends (asymptotically) of the inputs and the network topology. It becomes independent of its initial conditions~\cite{Jaeger09}. 
These conditions were summarized in the \textit{Echo State Property (ESP)}~\cite{Jaeger01}.
%
In practice, the stability of the ESN is almost always ensured when the spectral radius of the reservoir matrix is less than $1$~\cite{Jaeger09,Rodan11}.
As a consequence, the reservoir weights are appropriately scaled in order to have a spectral radius less than $1$.
To scale the parameters is necessary to compute the spectral radius of the reservoir matrix.
The computation of the spectra requires an important computational effort~\cite{Ferreira2013}.
%
%
Some attempts to generate a procedure for initializing the RC models were introduced in~\cite{Ferreira2013,Luko10,Boedecker09,Schrauwen07, BasterCord11}.

\subsection{$L_2$-Boost Using the ESN Model for Time-series Processing Information}
\label{Methodology}

In this article, we investigate the performance of using $L_2$-Boost in temporal learning tasks, and we consider as \textit{weak learner predictors} a set of ESNs with random initialization.
%
%
%
Given an arbitrary parameter $M$ the procedure is as follows. We initialize an ESN in a random way. 
The initialization consists in selecting the size of the network as well as the pattern of connectivity.
We consider a reservoir with fixed sparse connections. 
We do not control  the spectrum norm of the reservoir weight. 
A guide about the initialization procedure can be seen from~\cite{MantasPracticalGuide12}. 
We expand the input information using the ESN reservoir given by the expression~(\ref{reservoirState}), thus we obtain $\State^{(t)}$, $\forall t$. 
%
%
Next, we apply Algorithm~\ref{L2Boost}. 
Finally, we obtain predictor $F^{(M)}(\cdot)$. 
The approach is summarized in Algorithm~\ref{L2Boost-ESN}.
In our experiments we use ridge linear regression for computing the readout weights $\wo$~\cite{Rodan11}.

\begin{algorithm}[h!t]
\caption{The $L_2$-Boost with the ESN model.}
   \label{L2Boost-ESN}
\begin{algorithmic}
	\STATE Initialize an ESN following the comments in Subsection~\ref{Methodology};
	\STATE Compute the temporal expansion of $\cal{L}$ using~(\ref{reservoirState});
	\STATE Generate the set $\{(\State^{(t)},\y^{(t)}), \forall t\}$;
	\STATE Apply the Algorithm~\ref{L2Boost};
	\STATE Return $\F^{(M)}(\cdot)$;
\end{algorithmic}
\end{algorithm}


In order to evaluate the performance of this procedure, we compare the reached accuracy of the $L_2$-Boost technique with a simple \textit{baseline} approach~\cite{Bengio2000}.
The baseline approach consists in combining $K$ single predictors (in our case the learning predictors are ESNs). 
We consider random initialized reservoirs, without control of the reservoir spectrum norm. 
In the baseline method, we train independently each of these single ESNs. 
The final prediction is the average among the single predictions.
%
%

For statistical comparisons between the methods we consider $K=30$.
We remark again that we do not scale the reservoir weights for obtaining the ESP. 
Even though some ESN models can present good accuracy, other ones can be weak predictors.  
%
%
Additionally, we compare our performances with the performance obtained when single ESNs with ``good'' tuning of the reservoir parameters are used.
For that, we use the results presented in the RC literature. 
\section{Empirical Results}
\label{Results}
We begin this section describing the benchmark problems.
Next, we specify the experimental setup. 
We concludes this section with an analysis of our empirical results.

\subsection{Description of the Benchmark Problems}
\label{BenchmarkDescription}
We use the following range of time-series benchmarks:
\begin{itemize}
\item Fixed $k$th order NARMA. This data set presents a high non-linearity and is widely used in the RC literature. 
We generate the NARMA serie following the description in~\cite{Rodan11,Sakyasingha13},
\begin{center} 
$b(t+1)=\alpha_1(t)+\alpha_2 b(t)\displaystyle{\sum_{i=0}^{k-1}b(t-i)}$\\
$+\alpha_3 s(t-(k-1))s(t)+\alpha_4,$
\end{center}
where $s(t)\sim Unif[0,0.5]$ and the constants values are shown in Table~\ref{NarmaParameter}.
In order to evaluate the memorization ability of the model, we consider two simulated series when $k=10$ and $k=30$.
\begin{table}[h]
  \begin{center}
    \begin{tabular}{*{5}{c}}
    \hline\hline
    $k$ &  $\alpha_1$ & $\alpha_2$ & $\alpha_3$ & $\alpha_4$ \\
		\hline
		$10$ & $0.3$ & $0.05$ & $1.5$ & $0.1$\\
    $30$ & $0.2$ & $0.004$ & $1.5$ & $0.001$\\
		\hline\hline
    \end{tabular}
  \end{center}
\caption{Parameters considered for the fixed $k$th order NARMA serie with $k=10$ and $k=30$.}
  \label{NarmaParameter}
\end{table}
The task consists to predict the value $y(t+1)$ based on the history of $y(t)$ up to time $t$.
We used the first $200$ samples as initial washout in both the training and test procedure.
The regularization parameter used was $0.00001$.
\item The Santa Fe Laser data set~\cite{SantaFeLaser}. 
It is an experimental data that contains the intensity pulsations of a real laser recorded by a LeCroy oscilloscope.
The data is a cross-cut through periodic to chaotic intensity laser pulsations. 
These pulsations more or less follow the theoretical Lorenz model of a two level system~\cite{SantaFeLaser}. 
In this problem, the task consists to predict the next measure $y(t+1)$, given the precedent values up to $t$.
The original data only consists of $1000$ measurements, we used for training $499$ samples and for testing $500$ samples. 
We used a washout of $10$ samples.
%
%
%
The regularization parameter $\gamma$ was $0.001$.
\item Henon Map data set.  It is a prototypical invertible map with chaotic solutions proposed in~\cite{Henon76}. 
The data is generated by
$$y{(t+1)} =1-1.4 ({y{(t)}})^2+0.3y{(t-1)}+z(t+1),$$
where the noise is $z(t)\sim\mathcal{N}(0,0.05)$. The data is normalized in $[0,1]$.
The goal is to predict the next value $y(t+1)$ with the past information up to $t$.
We considered a training data with $3995$ samples and a test data with $795$ samples.
%
%
%
The regularization parameter $\gamma$ used was $0.001$.
We use an initial washout composed by $100$ samples.
The network topology has $3$ input units set with the last two precedent $y(t)$ values and the noise at current time. 
\item Freedman's non linear time data set~\cite{TimeSeriesFreedman}. 
The data is generated by
$$y{(t+1)}=g(y{(t)}),$$
where:
$$g(x)=\left\{ 
\begin{array}{ll} 
2x,  & {\rm{if }} x \leq 0.5,\\
2-2x, & x>0.5. 
\end{array} 
\right.$$
We consider a very short data set. 
The length of the training data was $30$ and the test size was $19$.
The initial value is $y{(0)}=0.23719$. 
The initial washout considered only was of $3$ samples.
The network topology has only one input unit, several reservoir units and one output unit. 
The regularization parameter $\gamma$ used was $0.001$.
%
%
\end{itemize}
%

\subsection{Experimental Setup}
\label{ExperimentalSetup}
\begin{table}[h!t]
  \begin{center}
    \begin{tabular}{*{5}{c}}
    \hline\hline
     $\;$ & Initial & $\;$  & Train & Test \\
		DATA & Washout & $\gamma$ & samples & samples \\
		\hline
		$10$th NARMA & $200$ & $0.00001$ & $1400$ & $2400$ \\
    $30$th NARMA & $200$ & $0.00001$  & $1600$ & $2600$\\
		Santa Fe Laser & $10$ & $0.001$ & $499$ & $500$\\
    Henon Map & $100$ & $0.001$ & $3995$ & $795$ \\
		Freedman's & $3$ & $0.001$ &  $30$ & $19$\\
    \hline\hline
    \end{tabular}
  \end{center}
\caption{Parameter setting of the benchmark problems. In all cases, we initialize the input weights ($\wi$) using Uniform distribution in $[-0.2,0.2]$, and we initialize the reservoir weights with an Uniform distribution in $[-0.8,0.8]$.}
\label{ParameterSetting}
\end{table}

We summarize the setting of the main parameters related to the benchmark problems in Table~\ref{ParameterSetting}. 
The table presents the initial washout period, the regularization parameter ($\gamma$) of the linear ridge regression, and the number of train and test samples for each benchmark problem.
%

%
%
The benchmarks selected have been widely used in the RC literature~\cite{Rodan11,Jaeger01,Verstraeten07,BasterCord11}.
In all cases, we use the \textit{Normalized Mean Square Error (NMSE)}  as measure of accuracy model~\cite{Jaeger09}.
The learning method used for computing the output weight matrix $\wo$ was the offline ridge regression.
This algorithm has a regularization parameter $\gamma$ that we adjust it for each benchmark problem.
The pre-processing data step consisted in normalizing the patterns in the interval $[0,1]$
We investigated the algorithm performance for several reservoir sizes.
The range of the reservoir size values is specified for each benchmark problem.
The connection between the input and reservoir layer is fully connected with random weights in $[-0.2,0.2]$.
The reservoir matrix is initialized using Uniform distribution in~$[-0.8,0.8]$.

\subsection{Result Analysis}
\label{ResultAnalysis}
%
%
 
Table~\ref{Compar} shows results reported in the RC literature for these benchmarks when a single ESN model was used as model predictor.
Table~\ref{TableHenon} presents the train set accuracy reached on the Henon Map data set. The columns $2$, $3$ and $4$ show the NMSE reached with $L_2$-Boost with ESNs for $M$ epochs ($M=3$, $4$ and $5$), respectively. 
Column $5$ of Table~\ref{Compar} shows the accuracy of the baseline approach, that it averaging the prediction of $30$ ESNs. The columns of the table are written using a scientific notation.

Table~\ref{TableHenon} illustrates the accuracy of the models during the training. 
The NMSE corresponds to the training data of the Henon Map data set. 
We present this table in order to illustrate the tendency of overfitting of $L_2$-Boost with ESN.
The additive model converge very fast to the solution, for this reason the columns~$3$ and $4$ of Table~\ref{TableHenon} are very similar.
The model with larger $M$ performs better over the train data, but it has problems of generalization.
We found this characteristic in all benchmarks. 
%
As a consequence, we can affirm that the parameter $M$ has a relevant impact in the control of the overfitting phenomenon.
We can found a similar remarks for the $L_2$-Boost technique in non-temporal learning tasks~\cite{Buhlmann03}.

\begin{figure}[h!]
\begin{center}
\fbox{\includegraphics[angle=0,width=0.45\textwidth]{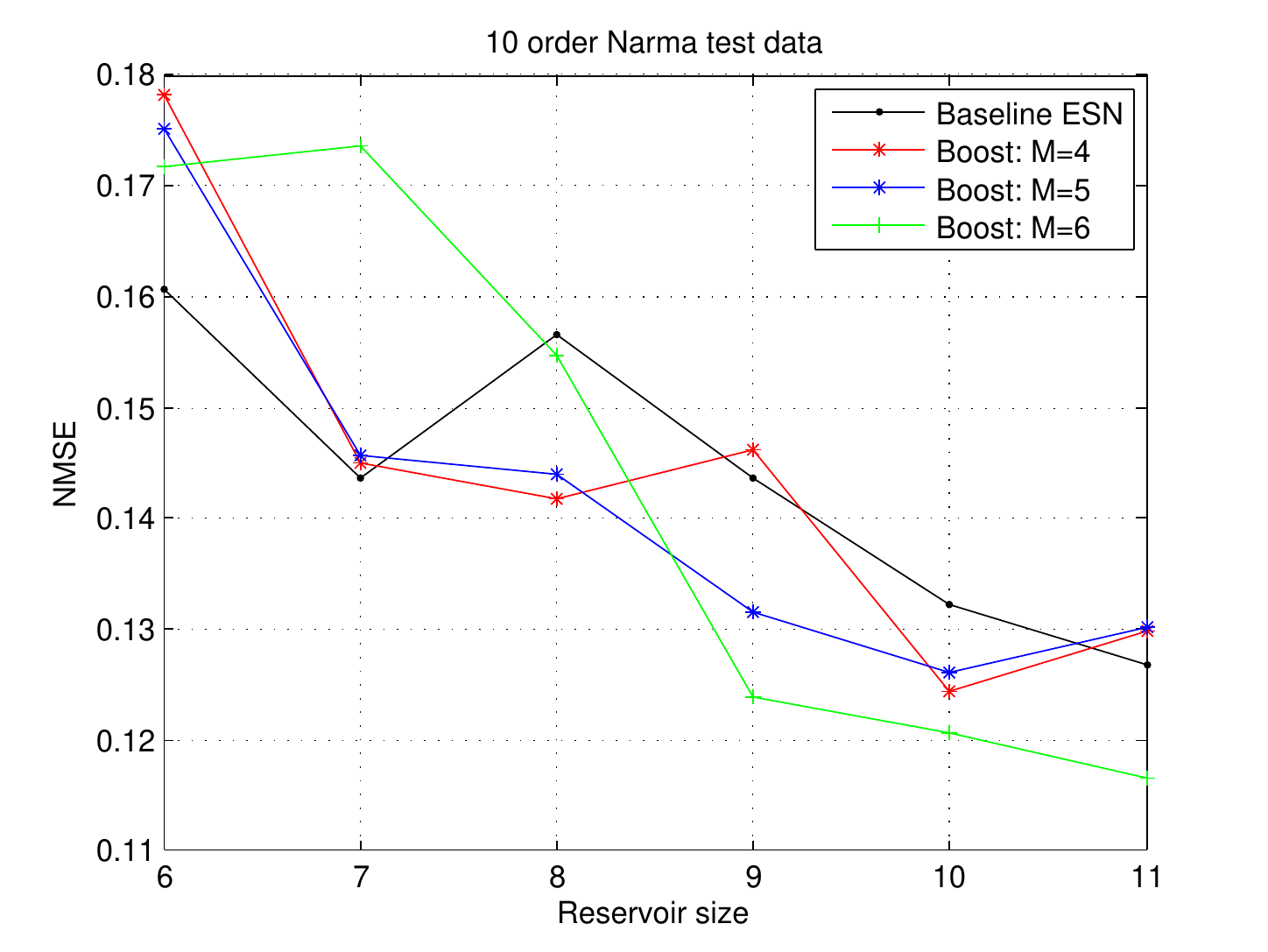}}
\caption{Test set accuracy reached for the $10$th order NARMA data set. 
The vertical axis of the graph shows the NMSE accuracy, and the horizontal axis presents some values of the reservoir size.  
We compare the accuracy of $L_2$-Boost: $M=4, 5$ and $6$ with the baseline approach  averaging $30$ ESNs.}
\label{Narma10Boost}
\end{center}
\end{figure}

\begin{figure}[h!]
\begin{center}
\fbox{\includegraphics[angle=0,width=0.45\textwidth]{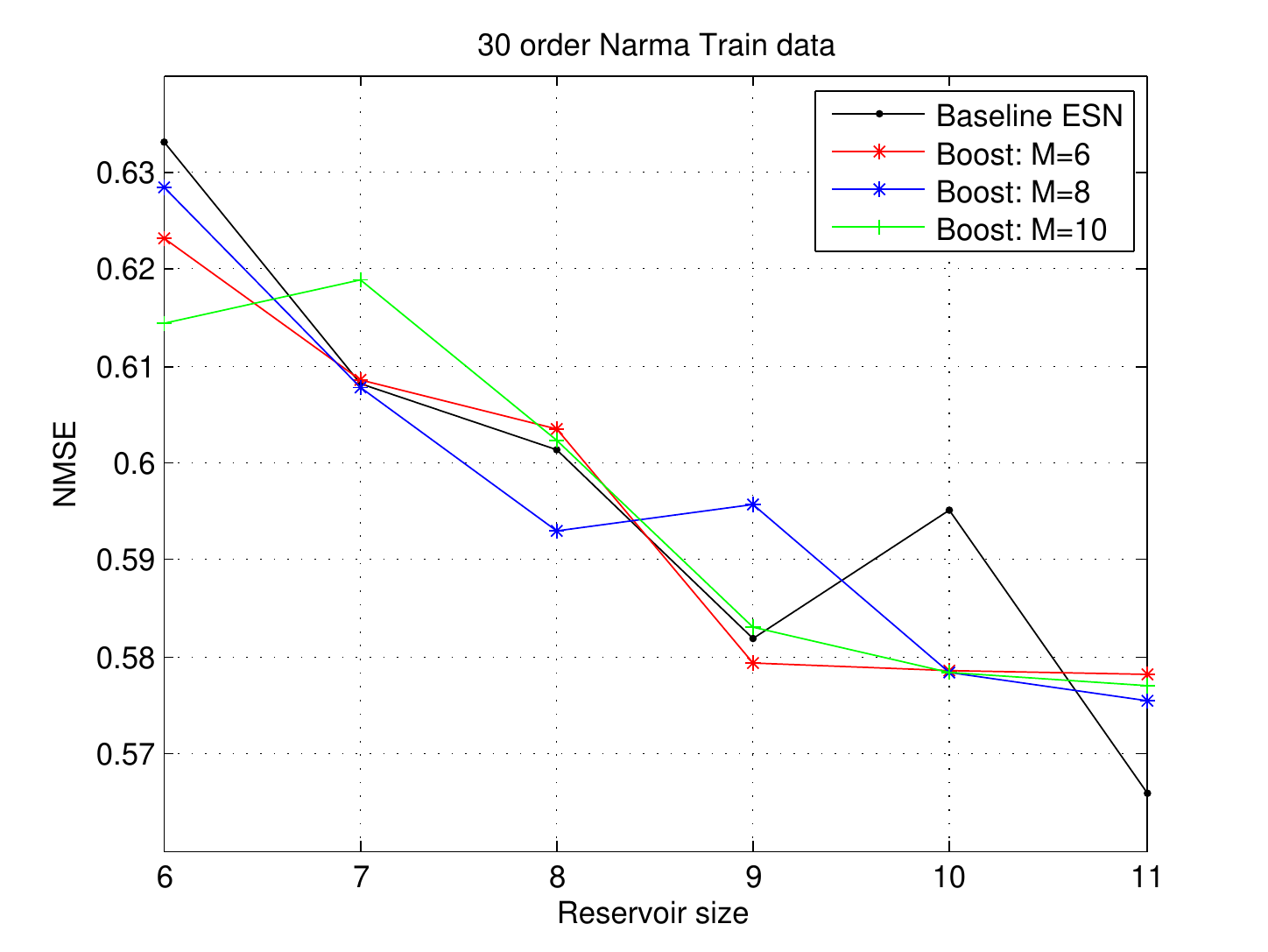}}
\caption{\label{Narma30TrainBoost}The accuracy reached on the training  set of the $30$th order NARMA data. 
The vertical axis of the graph shows the NMSE accuracy, and the horizontal axis presents some values of the reservoir size. 
We compare the accuracy of $L_2$-Boost: $M=6, 8$ and $10$ with the baseline approach  averaging $30$ ESNs.}
\end{center}
\end{figure}

\begin{figure}[h!]
\begin{center}
\fbox{\includegraphics[angle=0,width=0.45\textwidth]{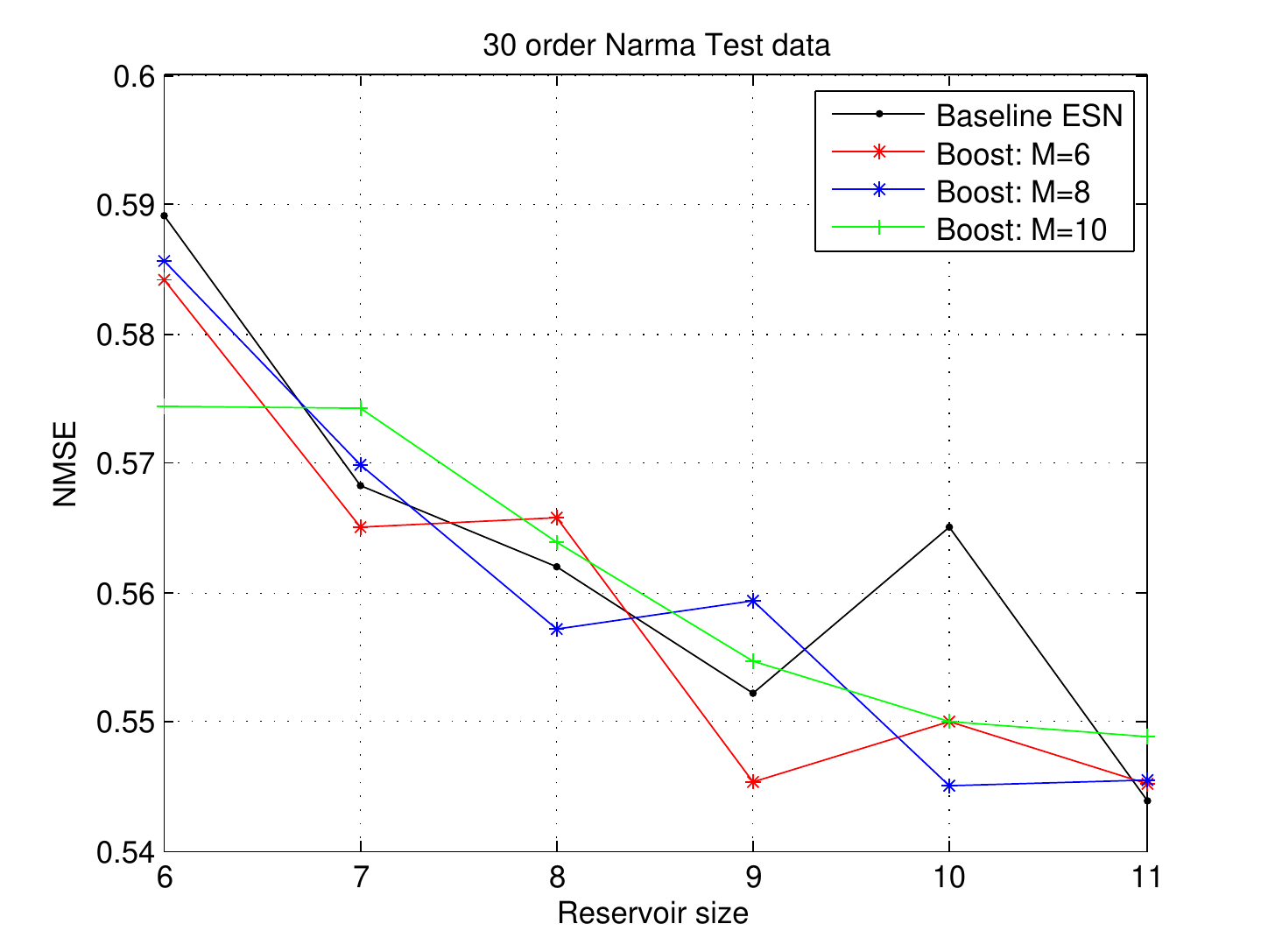}}
\caption{\label{Narma30Boost}The accuracy reached for the $30$th order NARMA data set. 
The vertical axis of the graph shows the NMSE accuracy, and the horizontal axis presents some values of the reservoir size. 
We compare the accuracy of $L_2$-Boost: $M=6, 8$ and $10$ with the baseline approach  averaging $30$ ESNs.}
\end{center}
\end{figure}

Figure~\ref{Narma30TrainBoost} illustrates the NMSE reached according the reservoir size for different $M$ values for the $30$th NARMA data set.
This figure shows the training error, we can see the evolution of the NMSE versus the size of the reservoir. We present few values of reservoir size between $6$ till $11$.
Figures~\ref{Narma10Boost} and~\ref{Narma30Boost} show the NMSE of the test data versus the reservoir size for the $10$th and $30$th order NARMA data set, respectively.
These figures show $4$ curves, the black one (with points represented by dots) corresponds to the baseline method which combines several single ESNs.
The other curves correspond to the $L_2$-Boost-ESN with different number of epochs $M=6, 8$ and $M=10$.
We can not affirm that the procedure of $L_2$-Boost with single weak ESNs performs better than \textit{optimal} single ESNs. The accuracy it is also of the same order that results presented in the RC literature using a single well-initialized ESN~\cite{Rodan11, BasterCord11}.
%

\begin{figure}[h!]
\begin{center}
\fbox{\includegraphics[angle=0,width=0.45\textwidth]{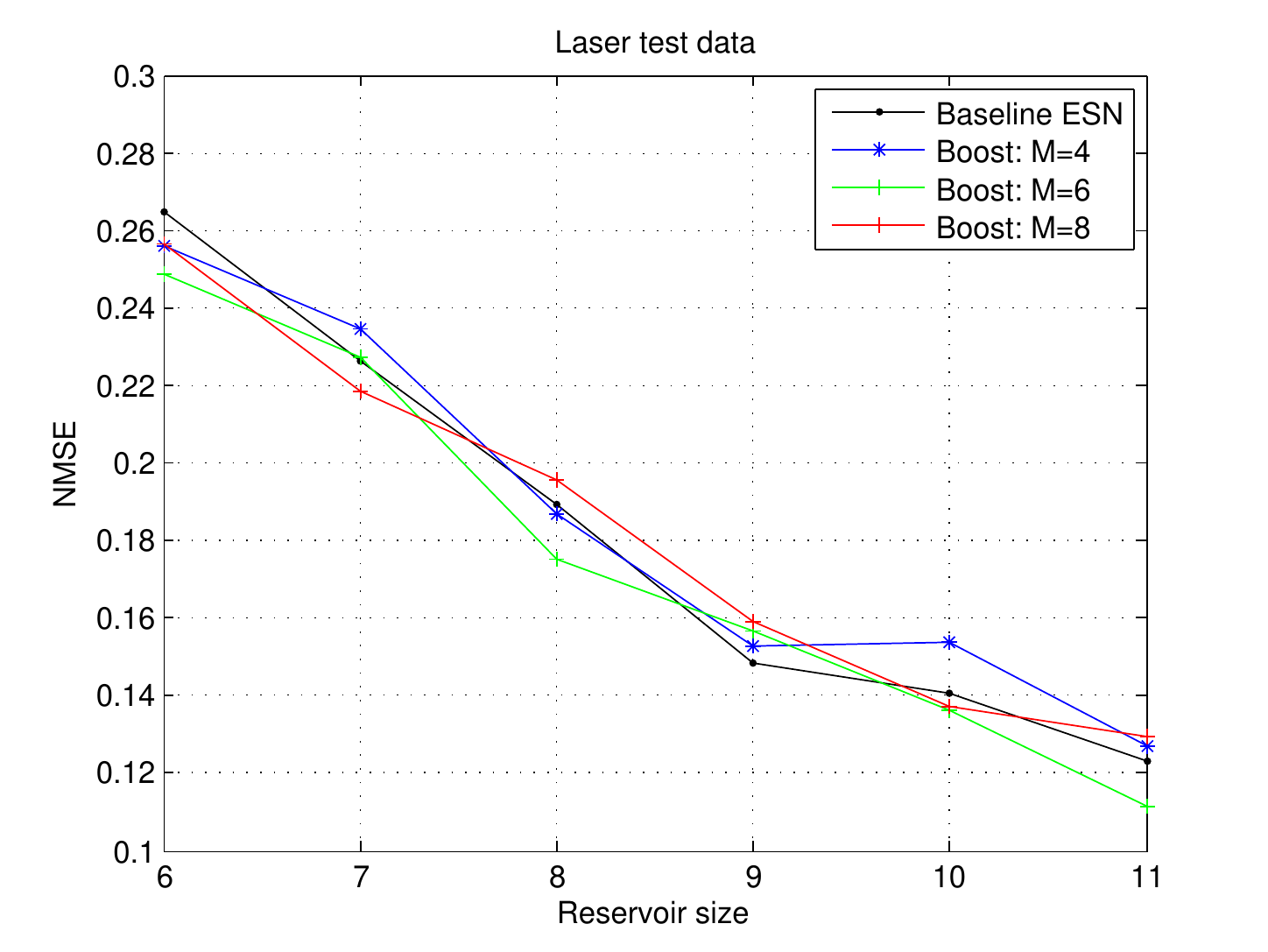}}
\caption{Test set accuracy of the Santa Fe Laser data. 
The vertical axis of the graph shows the NMSE accuracy, and the horizontal axis presents some values of the reservoir size.  
We compare the accuracy of $L_2$-Boost: $M=4, 6$ and $8$ with a baseline approach averaging $30$ ESNs.}
\label{LaserBoost}
\end{center}
\end{figure}

\begin{figure}[h!]
\begin{center}
\fbox{\includegraphics[angle=0,width=0.45\textwidth]{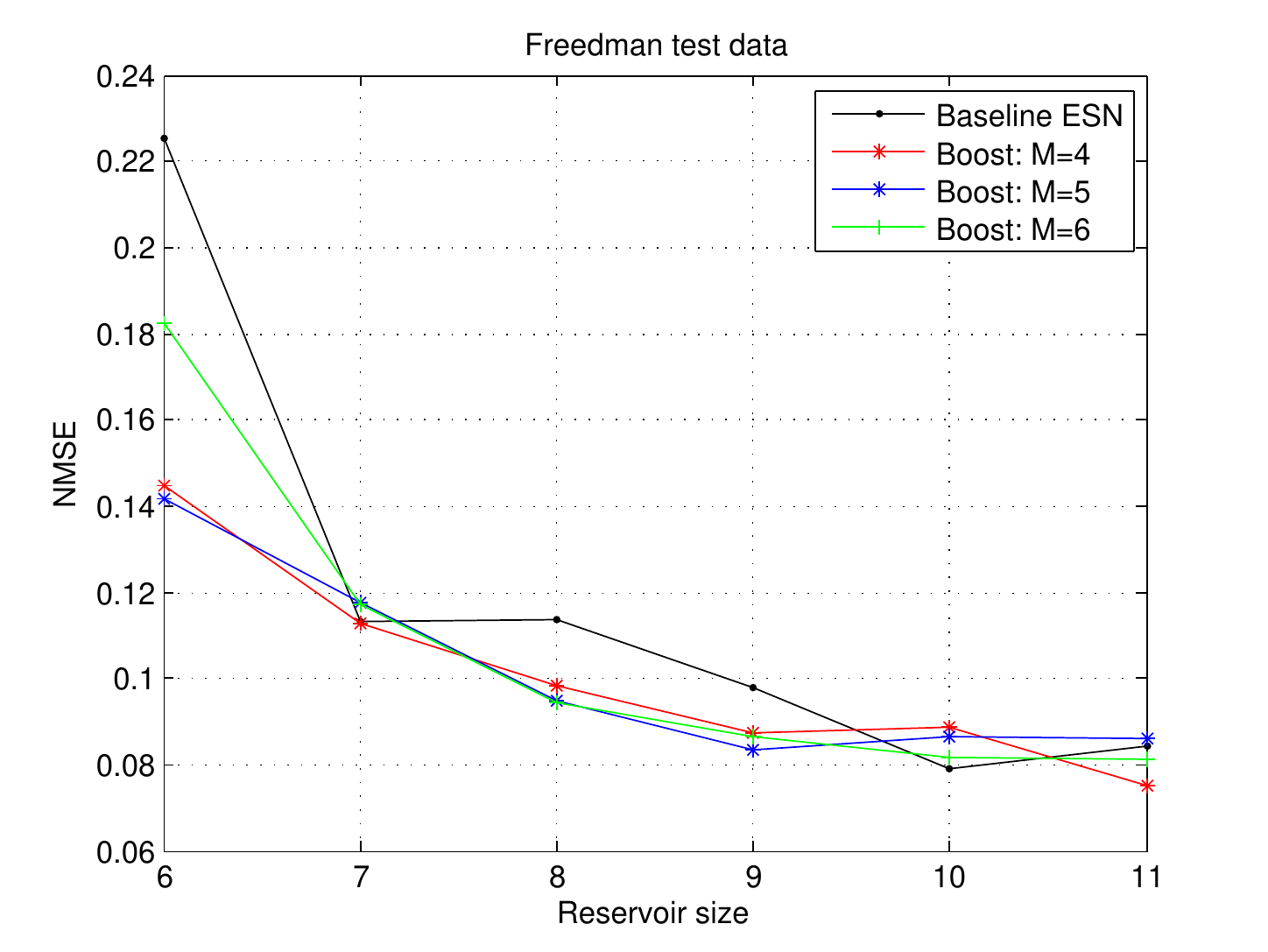}}
\caption{Test set performance of the Freedman data set. Size of the reservoir versus the NMSE accuracy. 
We compare the accuracy of $L_2$-Boost: $M=4, 5$ and $6$ with a baseline approach averaging $30$ ESNs.}
\label{FreedmanBoost}
\end{center}
\end{figure}

Figure~\ref{LaserBoost} illustrates the evolution of the NMSE for the reservoir size of the test data of the Santa Fe Laser benchmark. The error was computed for the $L_2$-Boost with ESNs for $M=4,6$ and $M=8$ and the baseline approach averaging $30$ ESNs.
Figure~\ref{FreedmanBoost} shows the accuracy reached for the models on the Freedman test data set.
The graphic shows the evolution of the $L_2$-Boost with ESNs for $M=4,5$ and $M=5$ and the baseline approach.
In all graphics, we can see that when the reservoir increases its size the procedure $L_2$-Boost with ESNs and the baseline approach decrease their test error.
This behavior about the impact of the reservoir size on the accuracy of the model, also happens with single ESNs~\cite{Rodan11,BasterCord11,MantasPracticalGuide12}.
\begin{table}[h!]
\begin{center}
\begin{small}
\begin{sc}
\begin{tabular}{llll}
\hline\hline
Data &  Accuracy & $\Nx$ & Ref. \\
\hline
$10$th NARMA & $0.166$ (NMSE) & $50$ &~\cite{Rodan11}\\
$\;$&  $0.0425$ (NMSE) & $200$ & ~\cite{Rodan11}\\\hline
$30$th NARMA & $0.4542$ (NRMSE) &  $100$ &~\cite{Boedecker09}\\\hline
Santa Fe Laser & $0.0184$ (NMSE) & $50$ &~\cite{Rodan11}\\
$\;$ & $0.00819$ (NMSE) & $200$ & ~\cite{Rodan11}  \\\hline
Henon Map &  $0.00975$ (NMSE) & $50$ &~\cite{Rodan11}\\
$\;$ & $0.00868$ (NMSE) &  $200$ & ~\cite{Rodan11}\\
Freedman's & $0.0004302$ (MSE) & $40$ & ~\cite{BasterCord11}\\
\hline\hline
\end{tabular}
\end{sc}
\end{small}
\end{center}
\caption{\label{Compar} Accuracy of the ESN model for the benchmark problems. 
Second column shows the accuracy reached by the single ESN, third column refers the reservoir size and the last column shows a bibliographic reference.
In the case of the Freedman's non linear time data, the reservoir initialization was done using the Scale Invariant Map method~\cite{BasterCord11}, and the Mean Square Error (MSE) was the error measure.
In the case of $30$th NARMA, the authors initialize the reservoir using permutation matrices. The error measure was the Normalized Root Square Error (NRMSE)~\cite{Boedecker09}.
In the other benchmarks problems, the authors control some reservoir parameters such as: spectral radius and reservoir matrix density. 
}
\end{table}
\begin{table}[h!]
\begin{center}
\begin{small}
\begin{sc}
\begin{tabular}{ccccc}
\hline\hline
$\;$ & $M=3$  & $M=4$ & $M=5$ & Bas. ESN \\
$\Nx$ & (1.0e-12) & (1.0e-13)& (1.0e-13) & (1.0e-12) \\
\hline
6 &   0.297367  &  0.131036 & 0.131036 & 0.288720\\
7 &  0.191883 & 0.131036  & 0.131036 & 0.064412 \\
8 &   0.142289 & 0.131036 & 0.131036 &   0.572728\\ 
9 &   0.160910 &  0.131036 & 0.131036 &  0.182942\\
10&  0.203845 & 0.131036 & 0.131036&  0.278237 \\
11&    0.272649  &  0.131036&  0.131035&  0.524017\\
12 &  0.488265 &    0.131036& 0.131036 &  0.459002\\
\hline\hline
\end{tabular}
\end{sc}
\end{small}
\end{center}
\caption{\label{TableHenon}Train set performance of the Henon Map data set. First column indicates the number of neurons in the reservoir. The columns $2$, $3$ and $4$ show the NMSE obtained with $L_2$-Boost with $M$ epochs ($M=3$, $4$ and $5$), respectively. Column $5$ shows the accuracy of the baseline approach. That is the accuracy average among 30 ESN predictions. The columns are written using a scientific notation.
}
\end{table}

%
%
%

%
\section{Conclusions and Future Work}
%


At the beginning of the 2000s, an efficient technique to train and design a RNN was developed under the name of Echo State Network (ESN).
This approach overcome the limitations to train RNN using the gradient descent method.
The performance of an ESN is highly dependent on its parameters and pattern of connectivity of the hidden-hidden weights 
%
%
%
Besides, the network setting can be computational expensive, in particular to compute the spectral radius of the hidden-hidden weight matrix.

In this article, we investigated boosting ideas with ESNs, in order to built a robust new learning tool.
In particular, we studied the utilization of $L_2$-Boost with random initialized ESNs. 
%
%
We merge a set of \textit{weak} single ESNs.  
We call \textit{weak} ESNs because they are random initialized, and we do not use extra computational effort for tuning the initial hidden-hidden weights.
%

%

%
In spite of the realization of numerous tests, we can not affirm that $L_2$-Boost with ESNs performs better than a single \textit{well-initialized} ESN (according the results presented in the RC literature).
%
%
%
However, the main advantage of the $L_2$-Boost with \textit{weak} ESNs is that the procedure is automatic and does not require the computational effort of computing the spectra of the hidden-hidden weight matrix.
%
Additionally, the procedure has a control parameter for the overfitting phenomena.

In a future work we will test the model using another technique for decrease the generalization error, as well as on a more number of benchmark problems.
Additionally, we can test the approach using another supervised learning tool for the readout structure.

%
%

%

%

\section*{Acknowledgment}
This work was supported within the framework of the IT4Innovations Centre of Excellence project, reg. no. CZ.1.05/1.1.00/02.0070 supported by Operational Programme 'Research and Development for Innovations' funded by Structural Funds of the European Union and state budget of the Czech Republic. Additionally, this article has been elaborated in the framework of the project New creative teams in priorities of scientific research, reg. no. CZ.1.07/2.3.00/30.0055.

\bibliographystyle{IEEEtran}
\bibliography{refRnn}

\vspace{5mm}
%
%
%

\end{document}